\documentclass[conference]{IEEEtran}
\IEEEoverridecommandlockouts

\usepackage[pagebackref,breaklinks,colorlinks]{hyperref}
\hypersetup{
    colorlinks,
    linkcolor={red!90!black},
    citecolor={green!90!black},
    urlcolor={blue!30!black}
}
\usepackage{url}            
\usepackage{booktabs}       
\usepackage{amsfonts}       
\usepackage{xcolor}         
\usepackage{amssymb}
\usepackage{makecell}

\usepackage{pifont}

\usepackage{arydshln}
\usepackage{wrapfig}
\usepackage{multirow}

\usepackage{cite}
\usepackage{amsmath}
\usepackage{algorithmic}
\usepackage{graphicx}
\usepackage{textcomp}
\usepackage{xcolor}
\def\BibTeX{{\rm B\kern-.05em{\sc i\kern-.025em b}\kern-.08em
    T\kern-.1667em\lower.7ex\hbox{E}\kern-.125emX}}
\usepackage{subfigure}

\begin{document}

\title{V2SFlow: Video-to-Speech Generation with\\Speech Decomposition and Rectified Flow}

\author{\begin{tabular}{c}
\textit{Jeongsoo Choi$^{1*}$, Ji-Hoon Kim$^{1*}$, Jinyu Li$^2$, Joon Son Chung$^1$, Shujie Liu$^2$}\\
$^1$Korea Advanced Institute of Science and Technology,
$^2$Mircrosoft \\
\{jeongsoo.choi, jh.kim, joonson\}@kaist.ac.kr, \{jinyli, shujliu\}@microsoft.com
\end{tabular}
\thanks{$^*$These authors contributed equally to this work. 
This work was supported by the National Research Foundation of Korea grant (MSIT, RS-2023-00212845) and by Institute of Information \& communications Technology Planning \& Evaluation (IITP) grant (MSIT, RS-2024-00457882, National AI Research Lab Project) funded by the Korean government.}
}

\maketitle

\begin{abstract}
In this paper, we introduce V2SFlow, a novel Video-to-Speech (V2S) framework designed to generate natural and intelligible speech directly from silent talking face videos. While recent V2S systems have shown promising results on constrained datasets with limited speakers and vocabularies, their performance often degrades on real-world, unconstrained datasets due to the inherent variability and complexity of speech signals. To address these challenges, we decompose the speech signal into manageable subspaces (content, pitch, and speaker information), each representing distinct speech attributes, and predict them directly from the visual input. To generate coherent and realistic speech from these predicted attributes, we employ a rectified flow matching decoder built on a Transformer architecture, which models efficient probabilistic pathways from random noise to the target speech distribution. Extensive experiments demonstrate that V2SFlow significantly outperforms state-of-the-art methods, even surpassing the naturalness of ground truth utterances.
\end{abstract}


\begin{figure*}[ht]
    \centering
    \includegraphics[width=0.89\textwidth]{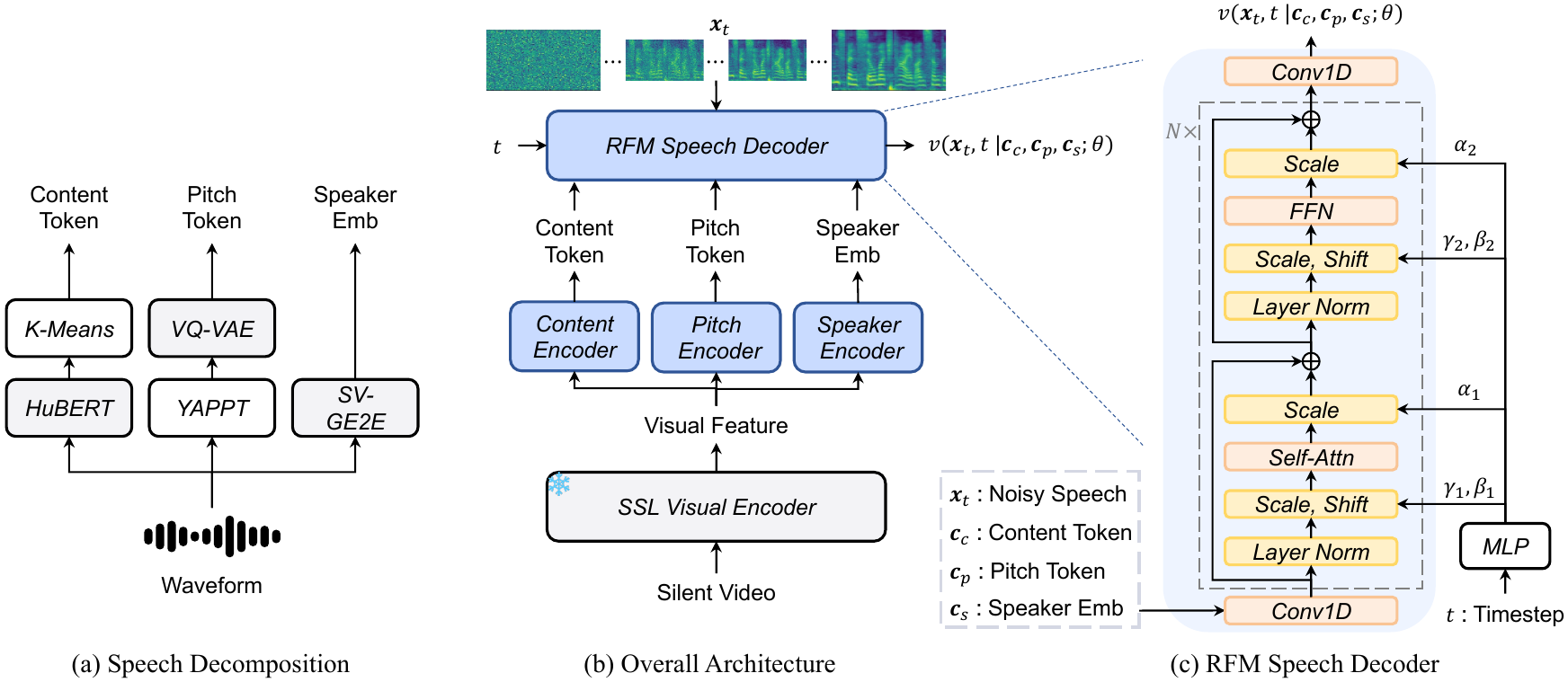}
    \caption{
        Overall framework of V2SFlow.
        In subfigure (a), SV-GE2E refers to the speaker verification model trained with GE2E loss.
        In subfigure (b), $\boldsymbol{x}_t$ denotes the intermediate noisy mel-spectrogram at timestep $t$, and $v$ represents the corresponding vector field.
        The parameters of the SSL video encoder are not updated during training.
        In subfigure (c), $\alpha$, $\beta$, and $\gamma$ represent the scale and shift parameters derived from the timestep.
    }
    \vspace{-3mm}
    \label{fig:main}
\end{figure*}

\section{Introduction}
Video-to-Speech (V2S) synthesis is an emerging research area aiming to generate natural-sounding speech solely from lip movements captured in silent video. This technology holds the potential to reconstruct human-like speech when the auditory signal is missing, opening up a variety of practical applications, such as enabling silent communication in secure environments and providing assistive technologies for individuals with aphonia. The advent of deep learning has revolutionized V2S synthesis by leveraging synchronized speech and lip movement sequences from video data for training, eliminating the need for additional annotations such as text transcriptions or speaker labels. This self-supervised nature significantly enhances the efficiency and scalability of V2S systems, paving the way for broader real-world applications.

Traditional V2S systems are typically trained on relatively small datasets with substantial constraints, such as controlled environments and a limited number of speakers~\cite{le2017generating, ephrat2017vid2speech, kumar2019lipper}. While these systems demonstrate promising results on such constrained datasets, their performance deteriorates when applied to real-world datasets with a larger number of speakers and a broader vocabulary. The primary challenge in achieving high-quality synthesis in real-world scenarios arises from the complex nature of speech signals, which encompass various acoustic components, including linguistic content and speaker characteristics. These components interact in nuanced ways, making it difficult for V2S systems to generalize effectively beyond the specific conditions of their training data.

To improve the generation quality in real-world scenarios, numerous studies have been conducted, broadly fall into two categories. The first focuses on modeling the inherent variability in speech to address the ambiguity between lip motions and the corresponding speech. To mitigate this ambiguity, several approaches incorporate text labels~\cite{kim2023liptospeech}, self-supervised speech units~\cite{hsu2023revise, choi2023intelligible, kim2024let, lei2024uni}, or extra lip-reading networks~\cite{hegde2023towards, yemini2024lipvoicer}. Other studies aim to clarify speaker identity by injecting speaker representations derived from either reference audio~\cite{mira22svts} or the input video itself~\cite{choi2023diffv2s}. The second research line seeks to better capture the complex dynamics of speech by leveraging advanced model structures and training algorithms, such as normalizing flows~\cite{he2022flow}, generative adversarial networks~\cite{kim2021lip, mira2022end, hegde2022lip}, and diffusion models~\cite{choi2023diffv2s,yemini2024lipvoicer}. Despite these advancements, current V2S systems still face challenges due to the inherent complexities of speech. Synthesized speech often suffers from artifacts, highlighting the need for more powerful architecture or training strategy to produce more natural outputs.

In this paper, we propose V2SFlow, a novel framework designed to generate natural and intelligible speech from silent video. To address the inherent ambiguity between lip movements and corresponding speech, we decompose speech into three fundamental and manageable subspaces: content, pitch, and speaker characteristics~\cite{polyak2021speech, choi2021neural}. For each speech sample, we extract content tokens, pitch tokens, and speaker embedding as representative features, which serve as prediction targets for training a video-to-speech model. Our model includes three specialized encoders, built upon a Self-Supervised Learning (SSL) visual encoder, to predict each of these decomposed attributes. With the predicted attributes, we employ a Rectified Flow Matching (RFM)-based decoder with Transformer backbone to reconstruct authentic speech. This approach combines the strengths of RFM~\cite{liu2023flow} and Diffusion Transformer (DiT)~\cite{peebles2023scalable}, enabling high-quality speech generation with a small number of sampling steps. Our extensive evaluations demonstrate the effectiveness of V2SFlow, even achieving superior naturalness compared to the ground truth speech. Audio samples can be found on our demo page\footnote{\url{https://mm.kaist.ac.kr/projects/V2SFlow}}.

\section{Method}

\begin{table*}[ht]
    \renewcommand{\arraystretch}{1.18}
    \renewcommand{\tabcolsep}{1.0mm}
    \caption{
        Video-to-speech synthesis performance comparisons on LRS3-TED and LRS2-BBC dataset.
        $\uparrow$: higher is better, $\downarrow$: lower is better.
    }
    \vspace{-2.5mm}
    \centering
    {
    \begin{tabular}{l ccccc ccccc}
        \Xhline{3\arrayrulewidth}
        \multirow{2.5}{*}{\textbf{Method}} & \multicolumn{5}{c}{\textbf{LRS3-TED}} & \multicolumn{5}{c}{\textbf{LRS2-BBC}} \\ \cmidrule(l{2pt}r{2pt}){2-6} \cmidrule(l{2pt}r{2pt}){7-11}
        & UTMOS$\uparrow$ & WER$\downarrow$ & SECS$\uparrow$ & MAE$_{F0}$$\downarrow$ & LSE-C$\uparrow$ & UTMOS$\uparrow$ & WER$\downarrow$ & SECS$\uparrow$ & MAE$_{F0}$$\downarrow$ & LSE-C$\uparrow$ \\ 
        \cmidrule(l{2pt}r{2pt}){1-1} \cmidrule(l{2pt}r{2pt}){2-6} \cmidrule(l{2pt}r{2pt}){7-11}
        Ground Truth & 3.519 & ~~2.5 & -- & -- & 7.63 & 3.017 & ~4.2 & -- & -- & 8.15  \\
        \cmidrule(l{2pt}r{2pt}){1-1} \cmidrule(l{2pt}r{2pt}){2-6} \cmidrule(l{2pt}r{2pt}){7-11}
        \multicolumn{5}{l}{\textcolor{gray}{\textit{with speaker embedding from audio}}} \\
        SVTS~\cite{mira22svts} & 1.256 & 78.0 & 0.557 & 0.389 & 6.04 & 1.349 & 80.9 & 0.593 & 0.374 & 7.91 \\
        Intelligible~\cite{choi2023intelligible} & 2.657 & 29.8 & 0.761 & 0.265 & \textbf{8.04} & 2.294 & 38.1 & 0.701 & 0.278 & 8.23 \\
        \textbf{V2SFlow-A} & \textbf{3.624} & \textbf{28.5} & \textbf{0.851} & \textbf{0.245} & 7.97 & \textbf{3.393} & \textbf{35.2} & \textbf{0.819} & \textbf{0.263} & \textbf{8.28}  \\ \cmidrule(l{2pt}r{2pt}){1-1} \cmidrule(l{2pt}r{2pt}){2-6} \cmidrule(l{2pt}r{2pt}){7-11}
        \multicolumn{5}{l}{\textcolor{gray}{\textit{with speaker embedding from video}}} \\
        DiffV2S~\cite{choi2023diffv2s} & 2.989 & 39.2 & 0.627 & 0.290 & 7.28  & 2.877 & 51.9 & 0.568 & 0.306 & 7.45  \\
        LTBS~\cite{kim2024let} & 2.428 & 79.7 & 0.607 & 0.289 & 7.84  & 2.319 & 86.4 & 0.534 & 0.306 & 7.74  \\
        \textbf{V2SFlow-V} & \textbf{3.780} & \textbf{28.5} & \textbf{0.664} & \textbf{0.251} & \textbf{8.09} & \textbf{3.648} & \textbf{35.6} & \textbf{0.581} & \textbf{0.275} & \textbf{8.39} \\
        \Xhline{3\arrayrulewidth}
    \end{tabular}
    }
    \label{table:1}
    \vspace{-5mm}
\end{table*}

\subsection{Speech Decomposition} 
Considering the complexity inherent in speech, we factorize it into distinct speech attributes and separately estimate them from silent video. This simplifies  the video-to-speech mapping, thereby facilitating more stable and effective training.

\noindent {\bf Content.}
Accurate linguistic content is crucial for generating intelligible speech, as it helps resolve ambiguities between lip movements and spoken words~\cite{kim2022distinguishing}. To extract this linguistic information without relying on additional text labels, we utilize the self-supervised speech model HuBERT~\cite{hsu2021hubert}. In particular, we employ HuBERT tokens discretized by K-means algorithm, as these tokens are known to capture detailed linguistic information independent of paralinguistic cues~\cite{lakhotia2021generative,kreuk2021textless,kim2024textless}.

\noindent {\bf Pitch.}
Pitch is a fundamental speech component, essential for conveying prosody and adding nuance beyond literal content. Building on recent works~\cite{polyak2021speech,polyak2021high,choi2024dddm}, we extract the pitch sequence from raw audio using \texttt{YAPPT}~\cite{kasi2002yet}, and then normalize it for each sentence to obtain distinctive pitch variations while reducing speaker information. This extracted pitch sequence is processed by a pre-trained VQ-VAE~\cite{van2017neural}, which encodes the sequence into a discrete latent space. The codebook from VQ-VAE provides quantized indices representing the pitch sequence, which we refer to as pitch tokens. These tokens serve as the representation of pitch in training our system, facilitating effective modeling and synthesis.

\noindent {\bf Speaker.}
We extract a global speaker embedding from a pre-trained speaker verification model optimized with GE2E loss~\cite{wan2018generalized}. This embedding captures speaker-specific characteristics across entire time dimension, helping the model reduce ambiguity caused by the varying traits of different speakers.

\subsection{Speech Attribute Estimation}
Given the decomposed speech attributes, we estimate each attribute directly from silent video based on recent findings that reveal unique associations between visual and acoustic features~\cite{hassid2022more,shi2022learning:avhubert,lee2024hear}. To do this, we utilize pre-trained AV-HuBERT (Large)~\cite{shi2022learning:avhubert} model, a powerful SSL audio-visual network capable of extracting both rich visual and linguistic information from the input video. Based on the visual features obtained from AV-HuBERT, we estimate each speech attribute through their respective encoders, as illustrated in Fig.~\ref{fig:main}(b).

We utilize Conformer~\cite{gulati2020conformer} for the content, pitch, and speaker encoders, as it is well-known for effectively capturing rich and contextualized features. Given the decomposed content tokens $\boldsymbol{c}_c$ from the target speech, as mentioned earlier, the content encoder is trained to estimate $\boldsymbol{c}_c$ from SSL visual features, and optimized by a frame-level cross-entropy (CE) loss with a label smoothing parameter set to $\alpha = 0.1$:
\begin{equation}
    \mathcal{L}_{c} = (1-\alpha)CE(\boldsymbol{c}_c, \hat{\boldsymbol{c}}_c) + \alpha CE(\boldsymbol{u}, \hat{\boldsymbol{c}}_c),
\end{equation}
where $\hat{\boldsymbol{c}}_c$ refers to the estimated content tokens, and $\boldsymbol{u}$ denotes a uniform distribution. Similar to the content encoder, the pitch encoder aims to predict the decomposed pitch tokens $\boldsymbol{c}_p$ and is trained with cross-entropy loss, also equipped with a label smoothing parameter $\alpha$ of 0.1.
\begin{equation}
    \mathcal{L}_{p} = (1-\alpha)CE(\boldsymbol{c}_p, \hat{\boldsymbol{c}}_p) + \alpha CE(\boldsymbol{u}, \hat{\boldsymbol{c}}_p),
\end{equation}
where $\hat{\boldsymbol{c}}_p$ refers to the generated pitch tokens. The speaker encoder takes SSL visual features 
as inputs and generates a single global speaker embedding by averaging the output features along the temporal dimension. To train this speaker encoder, we apply a cosine similarity loss defined as:
\begin{equation}
    \mathcal{L}_{s} = 1 -\text{cos}(\boldsymbol{c}_s, \hat{\boldsymbol{c}}_s),
\end{equation}
where $\boldsymbol{c}_s$ and $\hat{\boldsymbol{c}}_s$ represent the decomposed target and predicted speaker embeddings, respectively. Note that each encoder, including the subsequent speech decoder, is trained separately.

\subsection{Speech Decoder}
To reconstruct high-fidelity mel-spectrograms from the decomposed attributes, we introduce a Rectified Flow Matching (RFM)-based decoder with Transformer~\cite{vaswani2017attention} backbone. RFM~\cite{liu2023flow} estimates the vector field of the probability path from random noise to target data samples. Similar to diffusion-based models, it models the probability path between a tractable prior and target distributions. However, RFM aims to build straight paths between the prior distribution $\boldsymbol{x}_0 \sim p_0(\boldsymbol{x})$ and the target distribution $\boldsymbol{x}_1 \sim p_1(\boldsymbol{x})$, seeking to minimize the number of sampling steps. Given the condition $\boldsymbol{c}$ and an intermediate data sample $\boldsymbol{x}_t = (1-t)\boldsymbol{x}_0 + t\boldsymbol{x}_1$ at timestep $t \in [0, 1]$, RFM generative model $\theta$ is trained to estimate the vector field $u(\boldsymbol{x}_t, t | \boldsymbol{x}_1, \boldsymbol{c}) = \boldsymbol{x}_1 - \boldsymbol{x}_0$ with the RFM objective:
\begin{equation} \mathcal{L}_{RFM} = ||v(\boldsymbol{x}_t, t | \boldsymbol{c}; \theta) - (\boldsymbol{x}_1 - \boldsymbol{x}_0)||^2, 
\end{equation} 
where $v(\boldsymbol{x}_t, t | \boldsymbol{c}; \theta)$ is the estimated vector field.

In our case, we construct the condition $\boldsymbol{c}$ by concatenating all speech attributes along the channel dimension\footnote{We match the temporal lengths of speech attributes through linear interpolation. During training, the decomposed ground truth attributes are used, while during inference, the predicted values are utilized.}. Inspired by DiT~\cite{peebles2023scalable}, we utilize Transformer-based backbone and adaLN-Zero block to condition $\boldsymbol{t}$, facilitating effective generation. Moreover, we amplify the conditional sampling trajectory by adopting Classifier-Free Guidance (CFG)~\cite{ho2021classifier}. During training, we randomly drop all condition with a probability of 0.1. During inference, the speech decoder iteratively refines the mel-spectrogram, guiding the sampling trajectory away from unconditional flows. We use an Euler solver with CFG:
$\boldsymbol{x}_{t+\epsilon} = \boldsymbol{x}_t + \epsilon \{
\gamma v(\boldsymbol{x}_t,t|\boldsymbol{c};\theta) + (1-\gamma) v(\boldsymbol{x}_t,t|\varnothing ;\theta)
\},$
where $\epsilon$ and $\gamma$ denote the step size and guidance scale, respectively.

\section{Experimental Settings}
\subsection{Datasets}
We train our model on LRS3-TED~\cite{afouras2018lrs3}, which contains video clips featuring thousands of speakers and over 50,000 words. To enable unseen speaker evaluation, we adopt the unseen train-test split from SVTS~\cite{mira22svts}. Furthermore, to evaluate the generalizability in diverse environments, we use LRS2-BBC~\cite{chung2017lrs2} solely for inference. The video frame rate of both datasets is 25 fps, and the audio sample rate is 16 kHz.

\subsection{Evaluation Metrics}
We evaluate our method on each dataset using various metrics.
UTMOS~\cite{saeki2022utmos} estimates perceptual naturalness and Word Error Rate (WER) assesses the intelligibility of audio.
To calculate the WER, we use a pre-trained speech recognition model~\cite{ma2021end} to transcribe the audio clips and compare them to the ground truth text labels.
We measure Speaker Embedding Cosine Similarity (SECS), which assess the similarity between speaker representations\footnote{We use \texttt{Resemblyzer} python library to extract speaker representations.} of both the synthesized and target speech.
In addition, we obtain Mean Absolute Error of normalized F0 (MAE$_{F0}$) between the synthesized and target speech to assess the pitch accuracy, and also measure LSE-C~\cite{prajwal2020lip} using the pre-trained SyncNet~\cite{chung2017out} to evaluate lip synchronization accuracy.

\subsection{Implementation Details}
We crop, convert to grayscale, and apply data augmentation to the input video clips, as described in \cite{choi2023intelligible}. To crop the video, we detect the face using RetinaFace~\cite{deng2020retinaface} and extract facial landmarks using FAN~\cite{bulat2017far}. Meanwhile, the audio is converted into 80-bin mel-spectrogram with a hop size of 160 and a window size of 640. The resulting mel-spectrogram is then normalized based on the maximum and minimum values from the training dataset. We stack 20ms of the mel-spectrogram to obtain 50Hz feature.

We follow the Conformer block design from previous works~\cite{mira22svts,  choi2023intelligible} for our encoder architecture, and the speech decoder includes an 8-layer Transformer with a hidden dimension of 512 and 4 attention heads. Content, pitch, and speaker encoders are separately trained for 50k steps with a batch size of 144 seconds per GPU. We use 8 GPUs and adopt the same learning rate schedule as in \cite{choi2023intelligible}. The SSL visual encoder is kept frozen during training and the speech decoder is trained for 400k steps with a batch size of 384 seconds on a single GPU. We adopt logit-normal sampling~\cite{esser2024scaling} for sampling timestep ${t}$ and the number of sampling steps is set to 30. To convert mel-spectrogram to audible waveform, we employ HiFi-GAN~\cite{kong2020hifi}.

\subsection{Baseline Systems}
To investigate the impact of two different speaker embeddings—one derived from reference speech and the other predicted from silent video—we conduct experiments with two model variations: V2SFlow-A and V2SFlow-V. V2SFlow-A is compared to SVTS~\cite{mira22svts} and Intelligible~\cite{choi2023intelligible}, both of which use audio-driven speaker embeddings. V2SFlow-V is compared with two other baselines, DiffV2S~\cite{choi2023diffv2s} and LTBS~\cite{kim2024let}, which leverage video-driven speaker embeddings.

\begin{table}[t]
    \renewcommand{\arraystretch}{1.2}
    \renewcommand{\tabcolsep}{1.0mm}
    \caption{MOS results on LRS3-TED dataset.}
    \vspace{-0.2cm}
    \centering
    {
    \begin{tabular}{l ccc}
        \Xhline{3\arrayrulewidth}
        Method  & Naturalness$\uparrow$ & Intelligibility$\uparrow$ & Similarity$\uparrow$ \\ 
        \cmidrule(l{2pt}r{2pt}){1-1} \cmidrule(l{2pt}r{2pt}){2-4}
        Ground Truth &4.42$\pm$0.12 &4.88$\pm$0.06 &4.92$\pm$0.05 \\ \cmidrule(l{2pt}r{2pt}){1-1} \cmidrule(l{2pt}r{2pt}){2-4}
        SVTS \cite{mira22svts} & 1.03$\pm$0.03 & 1.49$\pm$0.12 &1.23$\pm$0.09 \\
        Intelligible \cite{choi2023intelligible} & 2.46$\pm$0.13 & 3.11$\pm$0.19 & 2.85$\pm$0.20 \\
        DiffV2S \cite{choi2023diffv2s} & 3.28$\pm$0.16 & 3.04$\pm$0.19 & 2.61$\pm$0.16 \\
        LTBS \cite{kim2024let} & 2.43$\pm$0.13 & 1.87$\pm$0.13 & 2.27$\pm$0.14 \\
        \cmidrule(l{2pt}r{2pt}){1-1} \cmidrule(l{2pt}r{2pt}){2-4}
        \textbf{V2SFlow-A} & 4.38$\pm$0.10 & 3.90$\pm$0.17 & {\bf 4.11$\pm$0.15} \\
        \textbf{V2SFlow-V} & {\bf 4.60$\pm$0.10} & {\bf 3.91$\pm$0.17} & 3.38$\pm$0.19 \\
        \Xhline{3\arrayrulewidth}
    \end{tabular}
    }
    \label{table:2}
    \vspace{-0.2cm}
\end{table}

\begin{table}[t]
    \renewcommand{\arraystretch}{1.2}
    \renewcommand{\tabcolsep}{0.5mm}
    \caption{Ablation study of Speech Decomposition on LRS3-TED dataset.}
    \vspace{-0.2cm}
    \centering
    {
    \begin{tabular}{l ccccc}
        \Xhline{3\arrayrulewidth}
        Method & UTMOS$\uparrow$ & WER$\downarrow$ & SECS$\uparrow$ & MAE$_{F0}\downarrow$ & LSE-C$\uparrow$ \\ \cmidrule(l{2pt}r{2pt}){1-1} \cmidrule(l{2pt}r{2pt}){2-6}
        V2SFlow-V & 3.780 & ~28.5 & 0.664 & 0.251 & 8.09 \\
        \cmidrule(l{2pt}r{2pt}){1-1} \cmidrule(l{2pt}r{2pt}){2-6}
        ~~\textit{w/o} $\textbf{c}_c$ & 3.271 & 106.4 & 0.643 & 0.259 & 2.81 \\
        ~~\textit{w/o} $\textbf{c}_p$ & 3.796 & ~28.5 & 0.663 & 0.255 & 8.11 \\
        ~~\textit{w/o} $\textbf{c}_s$ & 3.722 & ~28.7 & 0.587 & 0.252 & 7.99 \\
        ~~\textit{w/o} Decomp. & 3.534 & ~30.4 & 0.623 & 0.262 & 8.38  \\
        \Xhline{3\arrayrulewidth}
    \end{tabular}
    }
    \label{table:3}
    \vspace{-0.3cm}
\end{table}

\section{Experimental Results}
\subsection{Quality Comparison}
We evaluate the quality of V2SFlow against recent systems using both objective and subjective metrics, with the results presented in Table~\ref{table:1} and Table~\ref{table:2}, respectively. In Table~\ref{table:1}, V2SFlow consistently outperforms existing methods by a large margin. Our method even achieves a superior UTMOS compared to the ground truth, indicating that the synthesized speech from our method is more natural than the ground truth speech. The lowest WER indicates that V2SFlow synthesizes authentic speech with plausible linguistic content, while the best results in SECS and MAE$_{F0}$ demonstrate that the synthesized speech closely resembles the ground truth in terms of voice quality and pitch variations. 

In addition, we conduct subjective Mean Opinion Score (MOS) tests on the LRS3-TED dataset, where 30 domain experts rate the quality of 40 randomly selected audio clips on a scale from 1 to 5. As shown in Table~\ref{table:2}, our method significantly outperforms existing methods in terms of naturalness, intelligibility, and similarity. Consistent with the UTMOS results, our method achieves better naturalness than the ground truth speech. This implies that our method focuses solely on lip movement, resulting in clear speech that is independent of the background noise present in the ground truth speech.

\subsection{Ablation studies}
\subsubsection{Speech Attributes}
We explore the effect of decomposed speech attributes, with the results summarized in Table~\ref{table:3}. The findings demonstrate that each attribute makes an independent contribution to overall speech quality. Specifically, excluding $\boldsymbol{c}_c$ significantly degrades intelligibility and synchronization accuracy, as reflected by WER and LSE-C. The absence of $\boldsymbol{c}_p$ results in reduced pitch accuracy, as indicated by a higher MAE$_{F0}$. The speaker attribute $\boldsymbol{c}_s$ also plays a crucial role, with a noticeable drop in SECS when it is excluded. When we directly generate mel-spectrograms from visual features instead of estimating decomposed speech attributes, the overall quality is highly degraded, demonstrating the effectiveness of speech decomposition in V2S system.

\subsubsection{Speech Decoder}
Table~\ref{table:4} presents the results of our ablation study on the speech decoder. In the second row, we replace RFM with the Denoising Diffusion Implicit Model (DDIM)~\cite{song2020denoising}, maintaining the same number of sampling steps (30). The results consistently demonstrate RFM’s superiority over DDIM, with notable improvements in UTMOS, WER, SECS, and LSE-C metrics. Moreover, in the third row, we replace the adaLN-Zero conditioning with simple concatenation, as employed in DiffV2S~\cite{choi2023diffv2s}. This modification results in further performance degradation, particularly evident in the WER, SECS, and LSE-C scores, indicating the effectiveness of the adaLN-Zero conditioning method.

\subsubsection{Guidance scale}
To confirm the trade-off introduced by the guidance scale $\gamma$, we conduct a series of experiments, as shown in Table~\ref{table:5}. The first row demonstrates the clear advantage of using CFG, as $\gamma=1$—which corresponds to not using CFG—results in the worst performance. We find that $\gamma=2$ yields the best overall results, with only a slight trade-off in SECS and MAE$_{F0}$. 

\begin{table}[t]
    \renewcommand{\arraystretch}{1.2}
    \renewcommand{\tabcolsep}{0.5mm}
    \caption{Ablation study of Speech Decoder on LRS3-TED dataset.}
    \vspace{-0.2cm}
    \centering
    \resizebox{0.95\linewidth}{!}
    {
    \begin{tabular}{l ccccc}
        \Xhline{3\arrayrulewidth}
        Method & UTMOS$\uparrow$ & WER$\downarrow$ & SECS$\uparrow$ & MAE$_{F0}$$\downarrow$ & LSE-C$\uparrow$  \\ 
        \cmidrule(l{2pt}r{2pt}){1-1} \cmidrule(l{2pt}r{2pt}){2-6}
        V2SFlow-V & 3.780 & 28.5 & 0.664 & 0.251 & 8.09 \\
        ~~ \textit{w/} RFM $\rightarrow$ DDIM & 3.596 & 28.9 & 0.663 & 0.252 & 7.87  \\
        ~~~~~ \textit{w/} adaLN-Zero $\rightarrow$ Concat & 3.625 & 29.1 & 0.662 & 0.251 & 7.83  \\
        \Xhline{3\arrayrulewidth}
    \end{tabular}
    }
    \label{table:4}
    \vspace{-2mm}
\end{table}

\begin{table}[t]
    \renewcommand{\arraystretch}{1.2}
    \renewcommand{\tabcolsep}{1.0mm}
    \caption{Ablation study of guidance scale on LRS3-TED dataset.}
    \vspace{-0.2cm}
    \centering
    {
    \begin{tabular}{c ccccc}
        \Xhline{3\arrayrulewidth}
        $\gamma$ & UTMOS$\uparrow$ & WER$\downarrow$ & SECS$\uparrow$ & MAE$_{F0}$$\downarrow$ & LSE-C$\uparrow$  \\ 
        \cmidrule(l{2pt}r{2pt}){1-1} \cmidrule(l{2pt}r{2pt}){2-6}
        1  & 3.365 & 29.0 & 0.659 & {0.250} & 7.67  \\
        1.5  & 3.722 & 28.7 & 0.665 & 0.251 & 8.02  \\
        2  & {3.780} & {28.5} & {0.664} & 0.251 & {8.09}  \\
        4  & 3.541 & 28.7 & 0.655 & 0.251 & 8.03  \\
        \Xhline{3\arrayrulewidth}
    \end{tabular}
    }
    \label{table:5}
    \vspace{-3mm}
\end{table}

\section{Conclusion}
In this paper, we propose V2SFlow, which can effectively generate high-quality speech from silent videos. We focus on the intrinsic modelling complexity of speech, and address it by decomposing speech into fundamental speech attributes. Based on these attributes, we introduce a rectified flow-based speech decoder with Transformer architecture to generate high-fidelity speech with preserving acoustic and linguistic characteristics. Through extensive experiments, we have demonstrated the effectiveness of V2SFlow, and a comprehensive analysis further validates the reliability of our design.

\bibliographystyle{IEEEtran}
\bibliography{egbib}

\end{document}